\documentclass[a4paper,twocolumn]{esapub2005} 
\usepackage{times}
\usepackage{natbib}
\usepackage{url}

\usepackage{epstopdf}
\usepackage{url}
\usepackage[bottom]{footmisc}
\usepackage{makeidx}
\usepackage{graphicx}
\usepackage{multicol}
\usepackage{amsmath,bm,amsfonts,amssymb}
\usepackage{booktabs}
\usepackage{commath}
\usepackage{subfloat}
\usepackage{color,xcolor,ucs}
\usepackage[font=small,labelfont=bf]{caption}
\usepackage{tabularx}
\usepackage{float}

\usepackage{amsthm}
\usepackage{subfig}
\usepackage{wrapfig}
\usepackage{multirow}
\usepackage{listings}
\usepackage{lipsum}
\usepackage{textcomp}
\usepackage{algorithm}
\usepackage{algpseudocode}

\makeatletter
\algnewcommand{\LineComment}[1]{\Statex \hskip\ALG@thistlm \(\triangleright\) #1}
\algnewcommand{\LineCommentCont}[1]{\Statex \hskip\ALG@thistlm \parbox[t]{\linegoal}{\hangindent=1em\hangafter=1 $\triangleright$ #1}}
\makeatother

\newtheorem{defn}{Definition}

\title{A Framework for Task and Motion Planning based on Expanding AND/OR Graphs}
\author{Fulvio Mastrogiovanni}
\affil{University of Genoa, Italy, fulvio.mastrogiovanni@unige.it}
\author{Antony Thomas}
\affil{International Institute of Information Technology, Hyderabad, India, antony.thomas@iiit.ac.in}

\begin{document}


\maketitle

\begin{abstract}
Robot autonomy in space environments presents unique challenges, including high perception and motion uncertainty, strict kinematic constraints, and limited opportunities for human intervention. 
Therefore, \textit{Task and Motion Planning} (TMP) may be critical for autonomous servicing, surface operations, or even in-orbit missions, just to name a few, as it models tasks as discrete action sequencing integrated with continuous motion feasibility assessments. 
In this paper, we introduce a TMP framework based on expanding AND/OR graphs, referred to as \textsf{TMP-EAOG}, and demonstrate its adaptability to different scenarios. 
\textsf{TMP-EAOG} encodes task‐level abstractions within an AND/OR graph, which expands iteratively as the plan is executed, and performs in-the-loop motion planning assessments to ascertain their feasibility.
As a consequence, \textsf{TMP-EAOG} is characterised by the desirable properties of
(i) robustness to a certain degree of uncertainty, because AND/OR graph expansion can accommodate for  unpredictable information about the robot environment, 
(ii) controlled autonomy, since an AND/OR graph can be validated by human experts, and
(iii) bounded flexibility, in that unexpected events, including the assessment of unfeasible motions, can lead to different courses of action as alternative paths in the AND/OR graph. 
We evaluate \textsf{TMP-EAOG} on two benchmark domains.
We use a simulated mobile manipulator as a \textit{proxy} for space‐grade autonomous robots.
Our evaluation shows that \textsf{TMP-EAOG} can deal with a wide range of challenges in the benchmarks.
\end{abstract}

\section{Introduction}
\label{sec:introduction}

Robot autonomy is of the utmost importance for the next generation of space missions, from \textit{satellite assembly} and \textit{servicing} to \textit{planetary surface exploration} and \textit{in-situ resource utilisation}. 
It is foreseen that robots will operate with minimal human oversight, often under communication delays, constrained computational resources, limited sensing and unexpected changes in the environment.
Achieving mission goals will most likely demand not a \textit{single} action, but the careful \textit{sequencing} of multiple, interdependent operations, each to be validated for geometric feasibility and physical safety. 
For instance, grasping an instrument needed for a scientific experiment may require planning a collision-free trajectory in cluttered storage.
If the planned trajectory were obstructed, the robot would first identify the items originating the hindrance, and if possible rearrange surrounding, movable items in order to reach the instrument. 
This tight coupling between \textit{symbolic decision-making} and \textit{continuous geometric reasoning} is why \textit{Task and Motion Planning} (TMP) has received much attention in the past few years, and could be a critical capability in space robotics.  

TMP integrates task and motion planning in a unified framework, and has been studied extensively in manipulation~\cite{kaelbling2013IJRR, srivastava2014ICRA, dantam2016RSS}, object rearrangement~\cite{krontiris2014ICHR, krontiris2015RSS, karami2021AIIA}, navigation~\cite{lo2018AAMAS, thomas2021RAS}, and mobile manipulation~\cite{garrett2018IJRR, garrett2018IJRR1}. 
However, we argue that applications in space robotics amplify the challenges underlying these robot tasks.
TMP approaches must be robust to uncertainty, generalisable across diverse scenarios, and interpretable by human supervisors in safety-critical operations.  

While AND/OR graphs have been used to encode complex tasks by representing intermediate states as constraints, and (sequences of) actions as (hyper-)arcs among states, expanding AND/OR graphs allow for recursively applying the graph structure to tasks with non-deterministic (but foreseen) outcomes.
In this paper, we introduce a TMP framework based on expanding AND/OR graphs, referred to as \textsf{TMP-EAOG}, which is designed to integrate symbolic and geometric reasoning while maintaining properties desirable for (partially) autonomous space robots, namely robustness to uncertainty, controlled autonomy, and bounded flexibility. 

\begin{itemize}
\item
\textit{Robustness to uncertainty}.
It is expected that robots with advanced manipulation capabilities will operate in conditions characterised by limited sensing and unpredictable changes in the environment. 
These could require a robot to modify a course of action by selecting an alternative action sequence in the AND/OR graph, or to iteratively expand the graph to execute sequences of (previously unplanned) actions after having incorporated new information in the graph structure.
This makes \textsf{TMP-EAOG} resilient to incomplete or changing world models.
\item
\textit{Controlled autonomy}.
In space operations, fully autonomous decision-making is rarely acceptable without human oversight.
AND/OR graphs encode a task, that is, a whole decision process, into a legible and explainable structure.
\textsf{TMP-EAOG} enables mission planners to validate or constrain task branches, ensuring compliance with safety-critical rules and mission protocols. 
This balance between autonomy and human validation is key when communication delays make full teleoperation infeasible but accountability remains paramount.  
\item
\textit{Bounded flexibility}.
Flexibility is essential for autonomy, but in space it must be carefully bounded to prevent unsafe or overly resource-intensive behaviours. 
\textsf{TMP-EAOG} constrains flexibility to a well-defined set of allowed action sequences, which can be switched at run-time based on the reaching of previously defined mission states.
In any case, action sequences are within mission-approved task abstractions, as encoded in the AND/OR graph.
This ensures a certain degree of adaptability without compromising predictability or safety.  
\end{itemize}

We evaluate \textsf{TMP-EAOG} across two domains, inspired by widely recognised benchmarks in the TMP community~\cite{lagriffoul2018RAL}.
We consider the operations of a mobile manipulator, to be thought of as a proxy to space robots with advanced manipulation capabilities, and we focus on the \textit{action level} of \textsf{TMP-EAOG}.
In \textsf{BENCHMARK\_1}, inspired by the famous \textit{Towers of Hanoi} domain, the robot must sequence dexterous manipulation actions under dynamic reachability constraints.
This mirrors situations where a robot must handle tools or components around obstacles fixed to a lander deck, therefore requiring dynamic reordering of actions as reachability changes.  
In \textsf{BENCHMARK\_2}, which we refer to as the \textit{Habitat}, the robot must interleave symbolic state changes with geometric actions.
This benchmark reflects complex mixed-mode operations aboard a planetary habitat, where symbolic-level activities such as, for example, switching equipment to a ``ready'' state, or preparing consumables for a scientific experiment in the lab, must be combined with object manipulations.  
Beyond their academic value, we believe that these two benchmarks could serve as \textit{proxies} for space mission tasks, highlighting the capabilities enabled by frameworks like \textsf{TMP-EAOG} in addressing real-world challenges.

The paper is organised as follows.
Section \ref{sec:related_work} introduces relevant related work on Task and Motion Planning.
Section \ref{sec:preliminaries} describes useful concepts related to AND/OR graphs as a way to encode TMP tasks.
The overall behaviour of \textsf{TMP-EAOG} is introduced in Section \ref{sec:TMP_EAOG}.
Section \ref{sec:results} describes results in the two benchmarks.
Conclusions follow.

\section{Related Work}
\label{sec:related_work}

In the recent past, different TMP approaches have been developed for various, non space robotics related, application scenarios. 
The aSyMov planner described in~\cite{cambon2009IJRR} is shown to perform mobile manipulation tasks using a \textit{Forklift and Boxes} problem, where the forklifts move the boxes to specific locations. 
The classic \textit{Towers of Hanoi} problem is also accomplished, which involves object rearrangement and manipulation. 
In the work of Dornhege \textit{et al.}~\cite{dornhege2009SSRR, dornhege2009ICAPS}, mobile manipulation is the main focus with experiments involving rearrangement of different blocks. 
A TMP approach that incorporates estimation and perception uncertainties is developed in~\cite{kaelbling2013IJRR}.
The work therein focuses on a mobile manipulator performing household tasks. 
Manipulation in clutter or rearrangement planning is the focus of the works in~\cite{srivastava2014ICRA}. 
The planner in~\cite{lagriffoul2014IJRR} is shown to successfully execute different instances of \textit{pick and place}, \textit{filling a glass}, and \textit{stacking cups}. 
Maximizing the height of a physically stable construction from a collection of different objects is demonstrated in~\cite{toussaint2015IJCAI}. 
The IDTMP approach in~\cite{dantam2016RSS} is used to validate a \textit{Blocks World} problem where blocks are stacked in a specific order, as well as different instances of rearrangement planning. 
The FFRob planner~\cite{garrett2018IJRR} is evaluated on different rearrangement planning scenarios. 
The planner also evaluates experiments in a kitchen scenario where a robot is required to prepare a meal. 
Real-world kitchen manipulation tasks in the presence of environment uncertainty are demonstrated in~\cite{garrett2020ICRA}. 
\textit{Navigation Among Movable Obstacles} is the focus of the works in~\cite{stilman2005IJHR}. 
TMP in the context of robot navigation is explored in~\cite{lo2018AAMAS}. 
Both single and multi-robot TMP approaches for robot navigation incorporating state uncertainties are presented in~\cite{thomas2021RAS, thomas2020STAIRS}.  

The various approaches to TMP discussed above are based on different assumptions and focus on specific aspects of task and motion planning. 
Each planner mentioned previously may not possess inherent adaptability to effectively address the requirements of the various applications of TMP. 
In particular, many examples in the literature target structured domains (for example, kitchens, warehouses, or household settings), where sensing and control conditions are comparatively benign. 
In contrast, space robotics introduces harsher constraints: manipulation often occurs in cluttered but safety-critical environments, uncertainties cannot be easily reduced by additional sensing, and navigation is limited by energy and communication \textit{budgets}.  
For instance, a scenario resembling the \textit{Towers of Hanoi} benchmark can be found when a lander- or rover-mounted manipulator must handle tools around fixed obstacles on a spacecraft deck, where occlusions and reachability constraints change dynamically.  
Similarly, mixed symbolic–geometric tasks akin to the \textit{Habitat} benchmark are representative of scientific experiments in planetary habitats, where a robot may need to combine symbolic actions (for example, switching equipment states) with geometric manipulations of objects in a confined environment.  

In order to be able to accommodate these conditions, TMP planners need to incorporate properties such as robustness to uncertainty, generalisability across diverse scenarios, and explainability for mission validation.  
\textsf{TMP-EAOG} encodes the task-level abstractions efficiently and compactly within an AND/OR graph and in this way is able to adapt to different applications --- including those motivated by space operations --- simply by modifying the underlying graph structure. 

\section{Task and Motion Planning based on AND/OR Graphs}
\label{sec:preliminaries}

In this Section, we first introduce the basic terminology typically employed in TMP approaches, and then we show how AND/OR graphs provide a natural structure for combining symbolic reasoning and geometric feasibility assessment.

\subsection{Task and Motion Planning: Basic Concepts}

Task planning, or classical planning, is the process of finding a discrete sequence of actions from a current state to a desired goal state~\cite{ghallab2016book}. 
Formally:

\begin{defn}
A \textit{task domain} $\Omega$ can be represented as a state transition system $\Omega = \langle S, A, \gamma, s_0, S_g \rangle$, where $S$ is a finite set of states, $A$ is a finite set of actions, $\gamma : S \times A \rightarrow S$ is the transition function, $s_0 \in S$ is the start state, and $S_g \subseteq S$ is the set of goal states.
\end{defn}

As an example, for the operations of a planetary rover, $S$ may encode whether a sample container is empty or filled, $A$ may include actions such as \texttt{collect\_sample} or \texttt{stow}, and $S_g$ may represent the condition that a specific instrument has been successfully stored.

Motion planning, instead, finds a collision-free sequence of robot configurations that move the robot from a given start to a goal pose~\cite{latombe1991robot}. 
Formally:

\begin{defn}
A \textit{motion planning domain} is a tuple $M = \langle C, f, q_0, G \rangle$, where $C$ is the robot configuration space, $f=\{0,1\}$ is the collision indicator, $q_0 \in C$ is the initial configuration, and $G \subseteq C$ is the set of goal configurations.
\end{defn}

In a space environment, $C$ may correspond to the set of arm joint configurations of a manipulator mounted on a rover, and $G$ the set of feasible grasps on a tool fixed to the spacecraft deck.

TMP combines discrete task planning and continuous motion planning to facilitate efficient interaction between the two domains. 
Formally:

\begin{defn}
A \textit{task-motion planning domain} with task domain $\Omega$ and motion domain $M$ is a tuple $\Psi = \langle C, \Omega, \phi, \xi, q_0 \rangle$, where:
\end{defn}
\begin{itemize}
\item $\phi : S \rightarrow 2^C$ maps task states to regions of the configuration space;
\item $\xi : A \rightarrow 2^C$ maps discrete actions to corresponding motion plans.
\end{itemize}

This formalism captures the fact that a symbolic state (for example, \textit{sample container open}) is only meaningful if the robot can also realise a corresponding geometric configuration (for example, \textit{arm aligned with the container lid}).
We can therefore define the concept of \textit{task-motion planning problem}.

\begin{defn}
The task-motion planning problem for a task-motion domain $\Psi$ consists in finding a sequence of discrete actions $a_0, \ldots, a_n \in A$ such that $s_{i+1} = \gamma(s_i, a_i)$, $s_{n+1} \in S_g$, and a corresponding sequence of motion plans $\tau_0, \ldots, \tau_n$ such that for $i = 0, \ldots, n$, it holds that:
\begin{itemize}
\item $\tau_i(0) \in \phi(s_i) \ \textrm{and} \ \tau_i(1)  \in \phi(s_{i+1})$;
\item $\tau_{i+1}(0) = \tau_i(1)$;
\item $\tau_i \in \xi(a_i)$.
\end{itemize}
\end{defn}

\subsection{AND/OR Graphs for TMP}

An AND/OR graph~\cite{chang1971AI} provides a compact representation of such integrated reasoning.
It encodes alternative action sequences (OR) as well as compound requirements that must be satisfied simultaneously (AND). 
This makes AND/OR graphs particularly suited for planning in constrained and safety-critical environments because, as a matter fact, an AND/OR graphs encodes alternative logic rules that must be strictly satisfied.

\begin{defn}
An AND/OR graph $G$ is a directed graph $G = \langle N,H \rangle$, where $N$ is a set of nodes and $H$ a set of hyper-arcs.
\end{defn}

For our purposes, nodes represent world states (for example, \textit{instrument retrieved}), and hyper-arcs represent either alternative or combined actions (for example, \texttt{grasp\_from\_left} \textit{or} \texttt{grasp\_from\_right}) to achieve the same state. 
In the graph, the \textit{root} node represents the goal state, whereas \textit{leaf} nodes collectively represent the initial state, as if they were logic predicates.
The graph contains also a node representing \textit{failure}, that is, the impossibility of reaching the root node.
Starting from \textit{true} leaf nodes (that is, nodes which semantics correspond to a fact holding \textit{true} in the world), the graph is traversed along its possible paths to the root, and this induces the traversal of a set of intermediates states and the associated hyper-arcs, which encode the sequence of actions to carry out in order to reach the goal state.
For example, a rover endowed with a manipulator tasked with accessing a partly occluded rock sample could branch into two hyper-arcs, that is, one for \textit{moving the blocking rock} (in AND with \textit{grasping the target}), the other for \textit{repositioning the manipulator base}.

We also consider augmented AND/OR graphs and their recursive expansion to capture non-deterministic outcomes. This allows modelling cases where execution introduces new information, for example, detecting that a grasp is infeasible due to unexpected debris, and the graph must be expanded at run-time to remove clutter and try again.
This is done via the failure node.
When this node is reached, the whole AND/OR graphs is recursively attached to it, thereby requiring the process to \textit{reiterate}, with different values for leaf nodes, that is, different initial conditions. 

\begin{defn}
For an AND/OR graph $G=\langle N,H \rangle$, an augmented AND/OR graph $G^a = \langle N^a,H^a \rangle$ introduces a virtual node $n^v$ and hyper-arcs $H^v$ to represent possible expansions.
\label{def:augmented}
\end{defn}

\begin{defn}
An AND/OR graph network $\Gamma = \langle \mathcal{G}, T \rangle$ is a sequence of augmented graphs $G^a_i$ connected by transitions $t_i$ via virtual nodes, where each transition models the incorporation of new knowledge (for example, updating reachability after an attempted manipulation).
\label{def:aog_network}
\end{defn}

\textsf{TMP-EAOG} can be modelled as a graph network that grows and adapts as plan execution unfolds. 
For instance, inside a planetary habitat, a robot may attempt to place an object into a storage bay, only to discover that the bay is occupied, and therefore the original goal cannot be accomplished.
The graph then expands to include the subtask of clearing the bay, before returning to the original goal.

\section{Behaviour of \textsf{TMP-EAOG}}
\label{sec:TMP_EAOG}

In this Section we describe \textsf{TMP-EAOG}, which leverages AND/OR graph networks as described in Definition \ref{def:aog_network} to compactly encode task-level abstractions.
A comprehensive treatment of the formal aspects can be found in~\cite{Karamietal2025}.
In this paper, we emphasise the data processing flow, and its relevance for robots operating in space environments.

\begin{figure}[]
\centering
\includegraphics[width=0.45\textwidth]{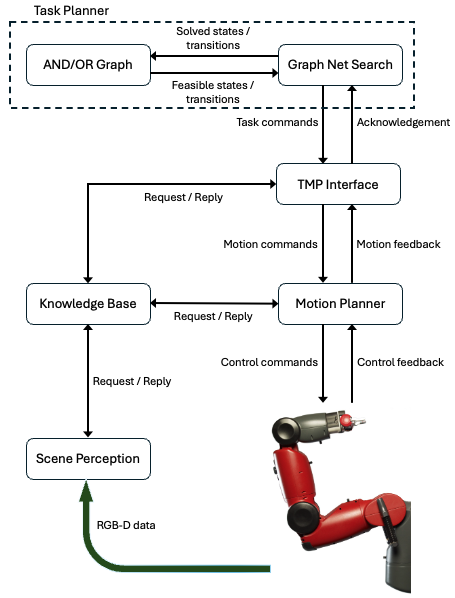}
\caption{System's architecture of \textsf{TMP-EAOG}, showing the interaction between perception modules, task-level reasoning, and motion feasibility evaluations.}
\label{fig:arch}
\end{figure}

Figure~\ref{fig:arch} illustrates the system's architecture of \textsf{TMP-EAOG}.
The process begins with the \textsf{Scene Perception} module, which collects information about the environment, for example, the location of tools, experiment samples, or structural elements.
Although, in the current implementation, mainly RGB-D data are used, it is noteworthy that in principle the module could be extended to process proprioception and touch data, as well as other sources of information.
All sensory data, aptly processed, are stored in the \textsf{Knowledge Base}, which maintains an up-to-date model of both the robot state and relevant traits of the environment.

The planning layer is composed of two key components, namely the \textsf{Task Planner} and the \textsf{Motion Planner} modules. 
The \textsf{Task Planner} module encapsulates the network of augmented \textit{AND/OR Graph} data structures, and the associated \textit{Graph Net Search} procedure to traverse the graph. 
This module generates candidate optimal transitions between states, and evaluates feasible sequences of actions as encoded in the hyper-arcs, receiving motion feasibility assessments as feedback. 
The \textsf{TMP Interface} module bridges the symbolic and geometric levels.
This module translates high-level, symbolic actions into geometric parameters (for example, end-effector poses, or rover base poses) and queries the \textsf{Motion Planner}. 
In this way, symbolic actions such as \texttt{grasp\_sample} or \texttt{relocate\_panel} are grounded into concrete robot configurations.
Although it is not the focus of this paper, it should be noted that \textsf{Task Planner} performs a series of in-the-loop simulations at the geometric level.
To this aim, the module first updates its internal representation of the robot environment via a query to \textsf{Knowledge Base}, and then carries out a series of motion planning simulations.
Examples of possible strategies can be found in~\cite{Thomasetal2021, Thomasetal2022, Thomasetal2023, Karamietal2025}. 

If the motion plan is feasible, the command is executed and an acknowledgement is returned to the \textsf{Task Planner}, which then updates the AND/OR graph with the new achieved state. 
Otherwise, if the motion plan is infeasible, the system backtracks and expands the graph, effectively exploring alternative action sequences (for example, clearing an obstacle before retrying the grasp). 
This mechanism makes \textsf{TMP-EAOG} particularly relevant in space robotics, where actions are executed in dynamic and uncertain environments and human intervention is limited.

\begin{algorithm}[t!]
\caption{\textsf{TMP-EAOG}}
\label{algo}
\footnotesize
\begin{algorithmic}[1]
\Require{
$\langle C, \Omega, \phi, \xi, q_0 \rangle$ : task-motion domain,
$\langle N^a,H^a \rangle$: augmented AND/OR graph}
\While{\texttt{Goal} == not-achieved}
\State{\texttt{AddNewGraph()}\label{AO}}
\State{\texttt{NextState} $\leftarrow$ \texttt{NextFeasibleStates()}\label{TP}}
\If{\texttt{NextState} == empty}
\State{\textbf{goto} line~\ref{AO}}
\Else
\State{\textbf{continue}}
\EndIf
\State{\texttt{OptimalState} $\leftarrow$ \texttt{NextOptimalState()}\label{NOP}}
\For{\texttt{(Tasks, Agents)} in \texttt{OptimalState}\label{TA}}
\LineComment Agents can be robot arms or different robots
\State{\texttt{RequestTMPI(Tasks, Agents)}\label{TMPI}}
\For{\texttt{Task} in \texttt{Tasks}}
\For{\texttt{Agent} in \texttt{Agents}}
\State{\texttt{RequestKB()}~\label{KB}}
\LineComment Call to \textsf{Knowledge Base}
\State{\texttt{RequestSP()}~\label{SP}}
\LineComment Call to \textsf{Scene Perception}
\State{\texttt{RequestMP()}}
\LineComment Call to \textsf{Motion Planner}
\EndFor
\EndFor
\EndFor
\If{\texttt{RequestMP()}}
\State{\texttt{MotionPlan} $\leftarrow$ \texttt{OptimalMotionPlan()}}
\If{\texttt{MotionPlan} == executed}
\State{\textbf{goto} line~\ref{TP}}
\Else
\State{\textbf{goto} line~\ref{AO}\label{retry}}
\EndIf
\Else
\State{go to line~\ref{NOP}}
\EndIf
\EndWhile
\State \Return{Task-Motion Plan}
\end{algorithmic}
\end{algorithm}

The overall \textsf{TMP-EAOG} behaviour is outlined in Algorithm~\ref{algo}. 
The Algorithm is presented qualitatively, as it aims to capture the iterative and feedback-driven nature of the framework rather than its implementation details. 
The process starts from the current state, in which the \textsf{Task Planner} module bootstraps the creation of the AND/OR graph by invoking the \texttt{AddNewGraph()} function (line~\ref{AO}). 
Subsequently, the planning phase proceeds by traversing the AND/OR graph from leaf nodes to the root, that is, from the current, initial state to the goal state (line~\ref{TP}).
Such traversal procedure, carried out in \textit{Graph Net Search}, has the cumulative effect of identifying the best path to the root node of the graph, that is, the best action sequence and the associated intermediate states.
In particular, \texttt{NextFeasibleStates()} determine the set of currently feasible states and the associated hyper-arcs, which induce the set of feasible actions to execute.
If there are no feasible states, the AND/OR graph must backtrack and expand.
Otherwise, the process identifies the best state among feasible ones using the \texttt{NextOptimalState()} subroutine. 
Once the \texttt{OptimalState} is determined, the associated \texttt{Tasks} are retrieved (line~\ref{TA}).
It is noteworthy that the graph can encode multi-agent behaviour, for example to model two separate robot arms; however, this is out of the scope of the present discussion and is reported only for the sake of completeness.
Both \texttt{Tasks} and \texttt{Agents} are shared with the \textsf{TMP Interface} module (line~\ref{TMPI}).

Then, the description of the relevant environment, the geometric coordinates of the objects therein, the location of the robot base, as well as the pose and joint configuration of robot manipulators are transmitted to \textsf{TMP Interface} through activating \textsf{Knowledge Base} and \textsf{Scene Perception} (lines~\ref{KB}-\ref{SP}). 
This exchange of information ensures that \textsf{TMP Interface} is supplied with the relevant and up-to-date data related to the involved components. 
\textsf{Motion Planner} is then executed to find a feasible trajectory to achieve the required goal configuration, as we mentioned above. 
Otherwise, if a feasible motion plan cannot be found, the AND/OR graph expands to a new graph to try out a new feasible state (line~\ref{NOP}), for example, moving an obstacle aside\footnote{
In general, a motion planner fails if no plan exists, or because the allotted planning time is insufficient.
In \textsf{TMP-EAOG}, we assume that sufficient time is allotted so that a motion planning failure implies the presence of objects obstructing the path to the target object or configuration.
If a target object is not graspable, or a configuration cannot be reached, then at least one impeding object should have to be rearranged to search for a new feasible motion plan.}.
For example, if a rover manipulator attempts to retrieve a sample and finds the trajectory blocked, the Algorithm expands the AND/OR graph to \textit{include the subtask} of moving the obstructing rock.
Finally, if the motion plan is successful and the goal is achieved, the algorithm terminates.     

\section{Results}
\label{sec:results}

We evaluate the capabilities of \textsf{TMP-EOAG} on two benchmarks.
\textsf{BENCHMARK\_1} implements the classical \textit{Tower of Hanoi} scenario, whereas \textsf{BENCHMARK\_2} describes a \textit{Habitat} environment where a robot is tasked of carrying out scientific experiments.

\subsection{\textsf{BENCHMARK\_1}: Tower of Hanoi}
\label{sec:benchmark_hanoi}

The first benchmark evaluates \textsf{TMP-EAOG} in a domain inspired by the classic \textit{Tower of Hanoi} problem, which is representative of sequencing dexterous manipulations under dynamic reachability constraints. 
In a space setting, this corresponds to situations in which a manipulator must handle tools or samples around fixed structures on a lander deck or inside a confined habitat, where occlusions and kinematic limits may render symbolic actions infeasible.

We simulated a variant of the Towers of Hanoi with a PR2 dual-arm manipulator. 
Disks of varying size were stacked on three rods arranged in a triangular configuration. 
At each step, only the top disk of a rod could be moved, and placement was constrained by disk size. 
The triangular arrangement meant that placing a disk on one rod could temporarily prevent reaching another rod, mimicking occlusions typical in cluttered space habitats.
PR2 is equipped with RGB–D sensing for object detection; its right and left arms were treated as separate \textit{agents} capable of performing handovers when a target rod was not reachable with a single arm.

\begin{figure}[t]
\centering
\includegraphics[width=0.45\textwidth]{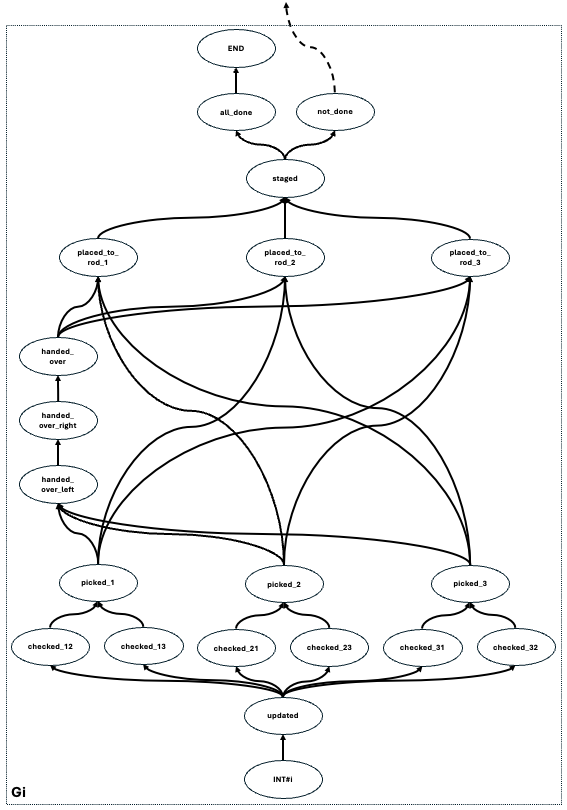}
\caption{An AND/OR graph for \textsf{BENCHMARK\_1}; hyper-arcs represent alternative task sequences.}
\label{fig:hanoi_graph}
\end{figure}

This domain highlights two critical features of space robotics autonomy: 
(i) the exponential growth of task spaces ($2^n-1$ moves for $n$ disks) and 
(ii) the frequent occurrence of symbolic actions that become infeasible at the geometric level, requiring graph expansion and re-planning. Figure~\ref{fig:hanoi_graph} shows an example AND/OR graph encoding the AND/OR graph.

We ran simulations with $n=3$ to $6$ disks. 
Table~\ref{tab:hanoi_av} reports average network depth $d$ (that is, the number of nested AND/OR graphs in expansion), task planning time $TP$, motion planning time for each arm, and number of motion planning attempts. 
As $n$ increases, the number of motion planning calls grows significantly, with noticeable imbalance between the two arms due to workspace geometry. 
Moreover, the observed network depth $d$ grows faster than the theoretical optimum $2^n-1$, reflecting the additional expansions required to resolve infeasible actions and kinematic constraints. 
For instance, with $n=3$, the optimal depth is $7$, but the average observed was $16.7$; with $n=6$, depth reached $140$. 
These results demonstrate how \textsf{TMP-EAOG} can capture non-monotonic rearrangements and dynamically reconfigure plans in the presence of occlusions and failures, properties critical for robust space autonomy.

\begin{table*}[t]
\centering
\scalebox{0.7}{
\begin{tabular}{ccccccc} 
\toprule
\textit{Objects}    & \textit{d}        & \textit{TP} [s]   & \textit{Right MP} [s] & \textit{Right attempts}    & \textit{Left MP} [s]  & \textit{Left attempts} \\
\hline
\hline
3                   & 16.7$\pm$3.5    & 0.78$\pm$0.01     & 105.7$\pm$18.1      & 76.1$\pm$0.7                & 111.4$\pm$12.2      & 88.1$\pm$4.4 \\
4                   & 31.0$\pm$8.2    & 3.27$\pm$0.37     & 245.6$\pm$47.9      & 200.2$\pm$20.1              & 186.7$\pm$15.8      & 129.3$\pm$0.3 \\
5                   & 61.0            & 15.2$\pm$0.6      & 458.2$\pm$1.9       & 442.7$\pm$20.3              & 409.8$\pm$14.5      & 315.9$\pm$16.3 \\
6                   & 140.0           & 59.7              & 1017.3              & 854.2                       & 678.7               & 475.5 \\
\bottomrule
\end{tabular}}
\caption{Average metrics for \textsf{BENCHMARK\_1}; increasing disks lead to deeper networks and more motion planning attempts, due to infeasible actions and kinematic limits.}
\label{tab:hanoi_av}
\end{table*}

\begin{figure*}[t!]
\centering
\subfloat[]{\includegraphics[width=3.9cm,height=2.9cm]{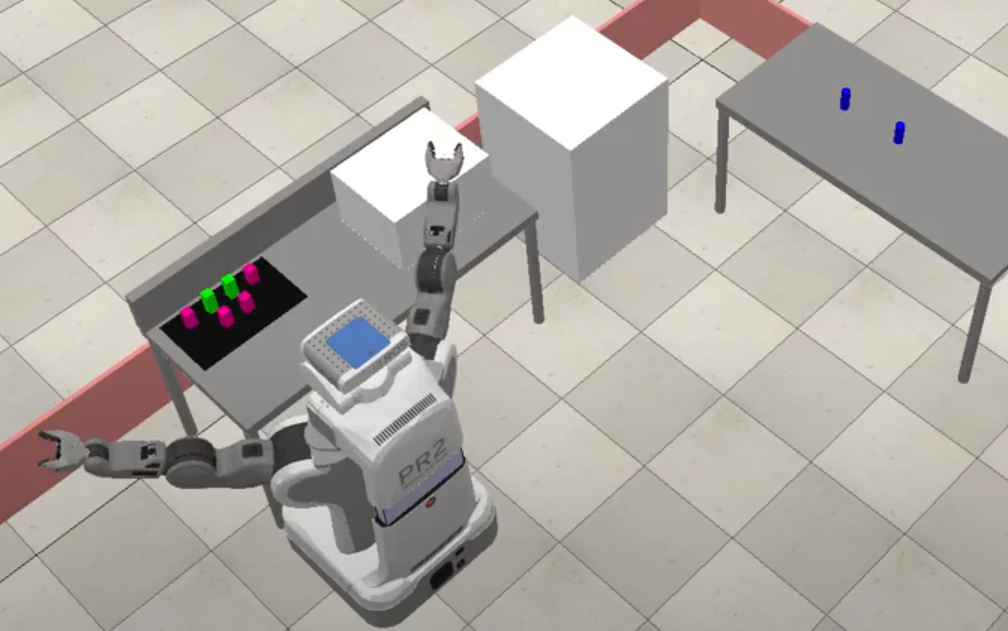}
}
\subfloat[]{\includegraphics[width=3.9cm,height=2.9cm]{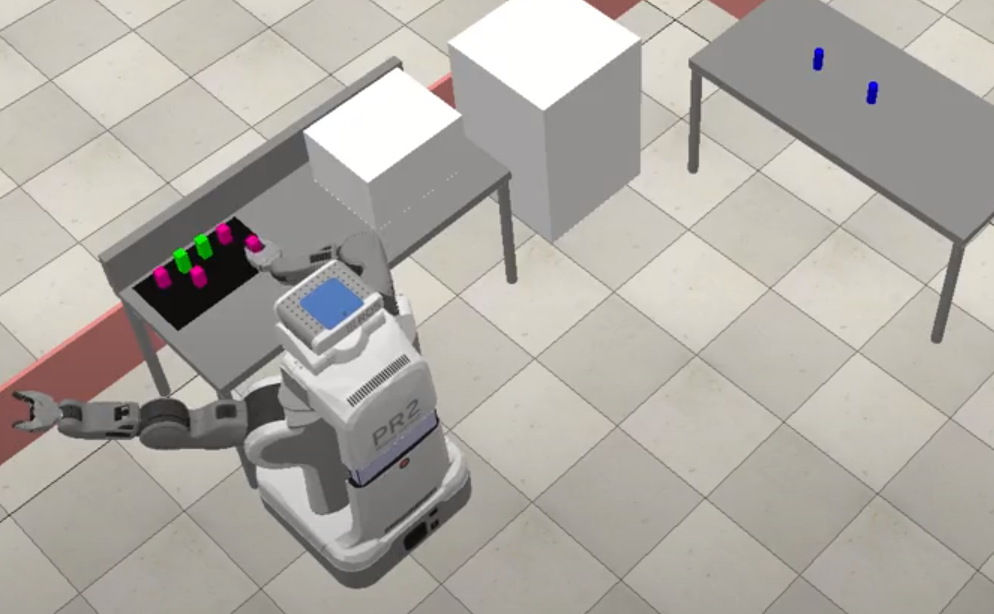}
}
\subfloat[]{\includegraphics[width=3.9cm,height=2.9cm]{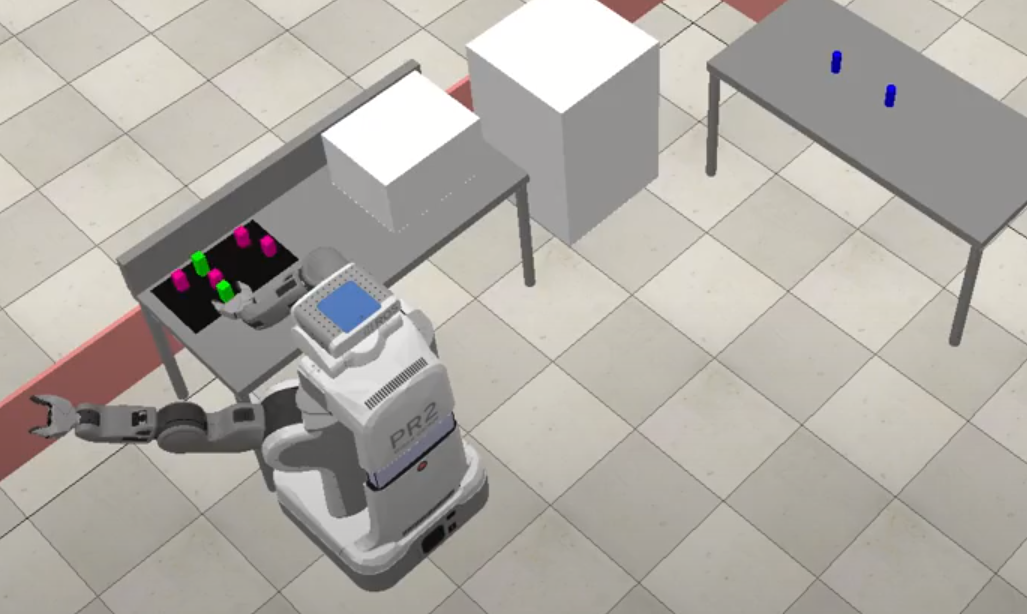}
}
\subfloat[]{\includegraphics[width=3.9cm,height=2.9cm]{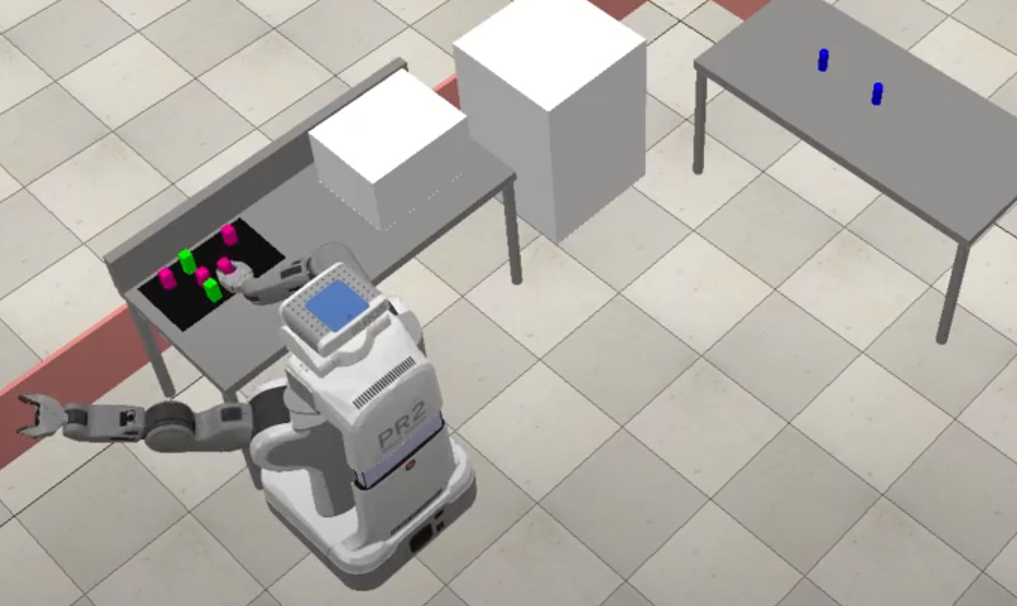}
}\\
\subfloat[]{\includegraphics[width=3.9cm,height=2.9cm]{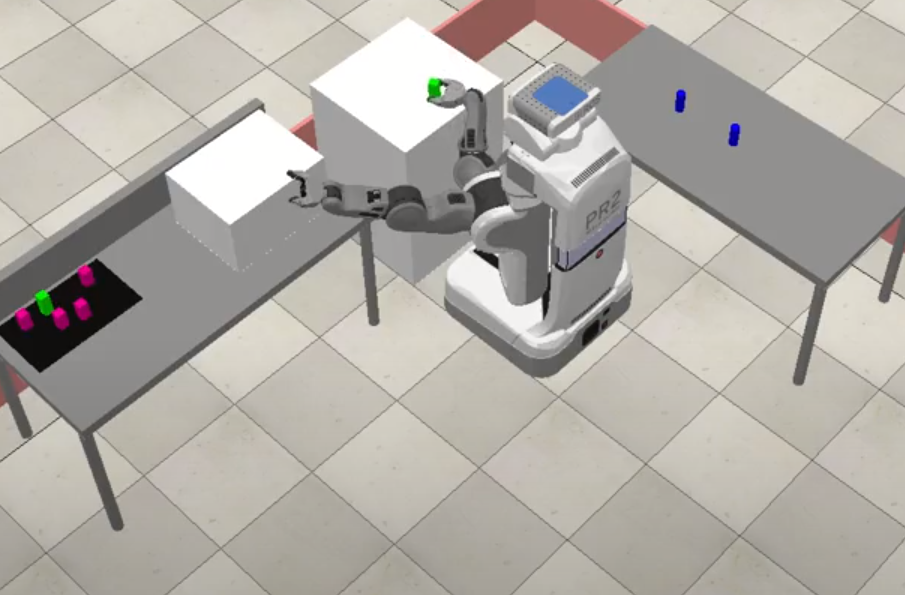}
}
\subfloat[]{\includegraphics[width=3.9cm,height=2.9cm]{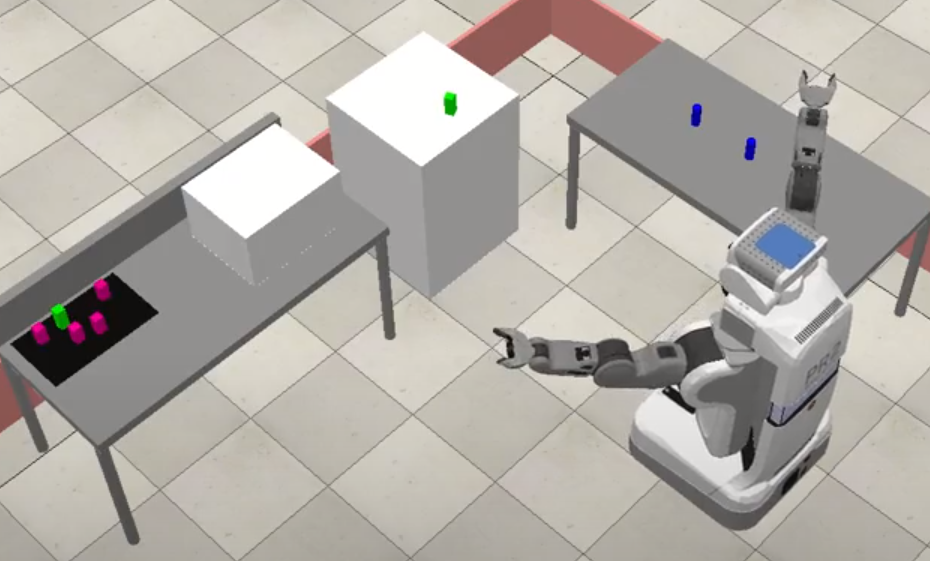}
}
\subfloat[]{\includegraphics[width=3.9cm,height=2.9cm]{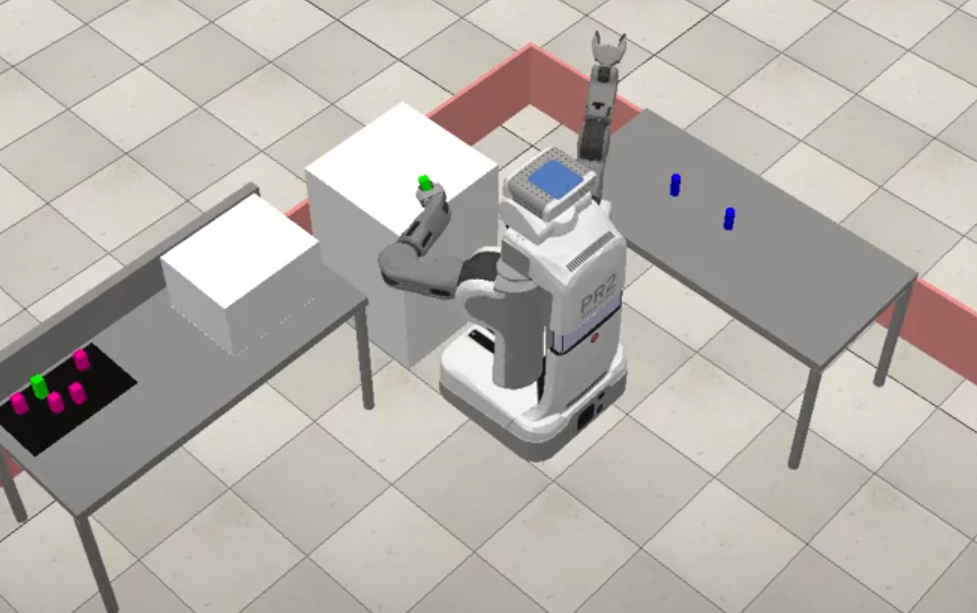}
}
\subfloat[]{\includegraphics[width=3.9cm,height=2.9cm]{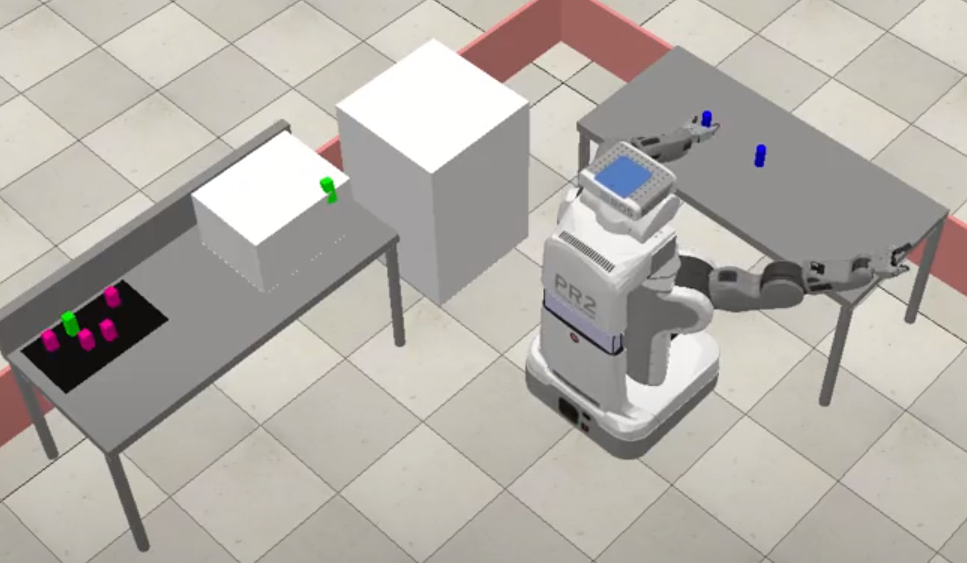}
}\\
\subfloat[]{\includegraphics[width=3.9cm,height=2.9cm]{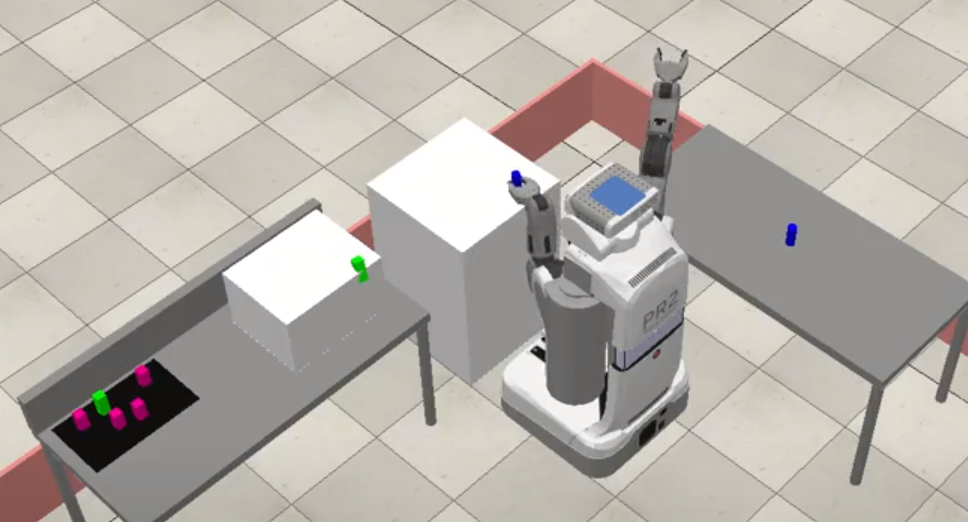}
}
\subfloat[]{\includegraphics[width=3.9cm,height=2.9cm]{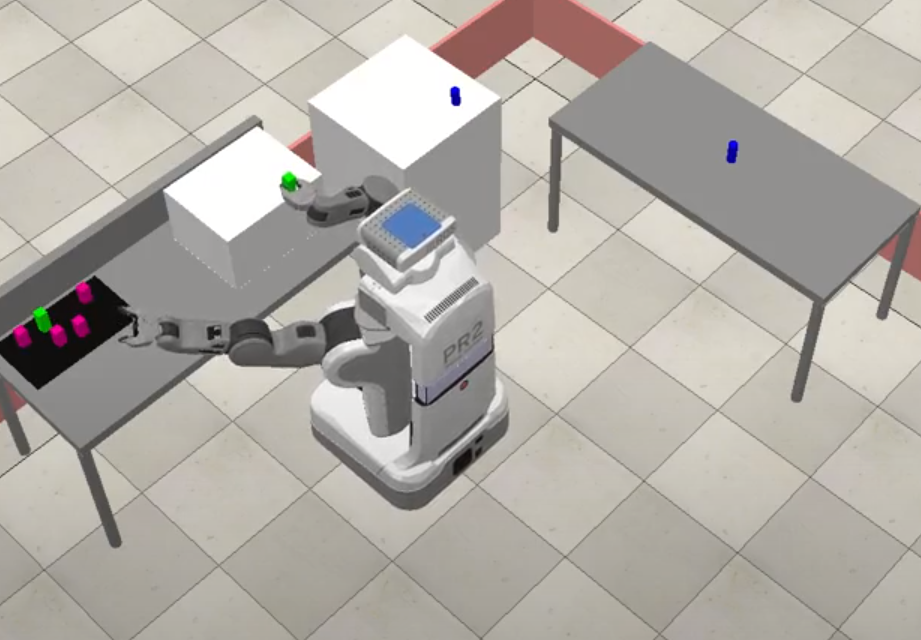}
}
\subfloat[]{\includegraphics[width=3.9cm,height=2.9cm]{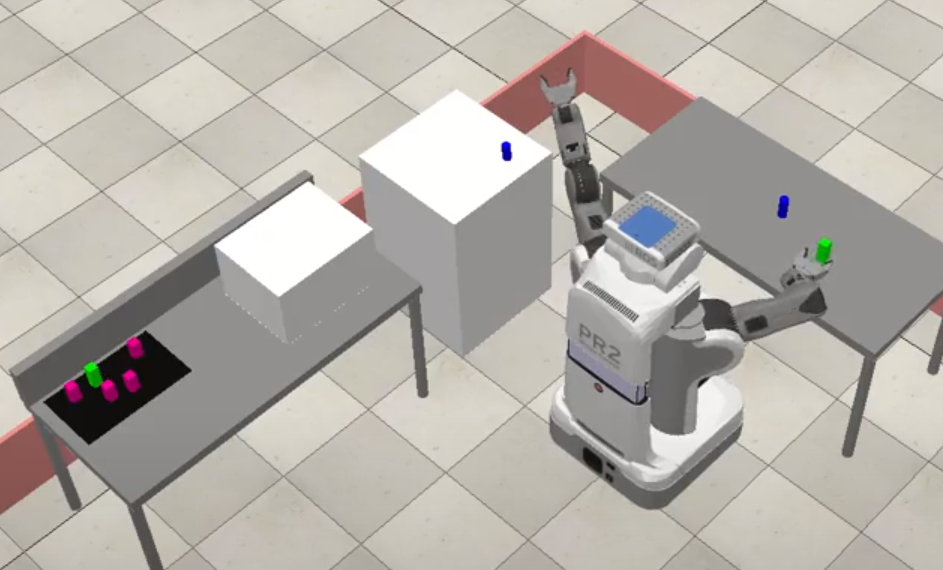}
}
\subfloat[]{\includegraphics[width=3.9cm,height=2.9cm]{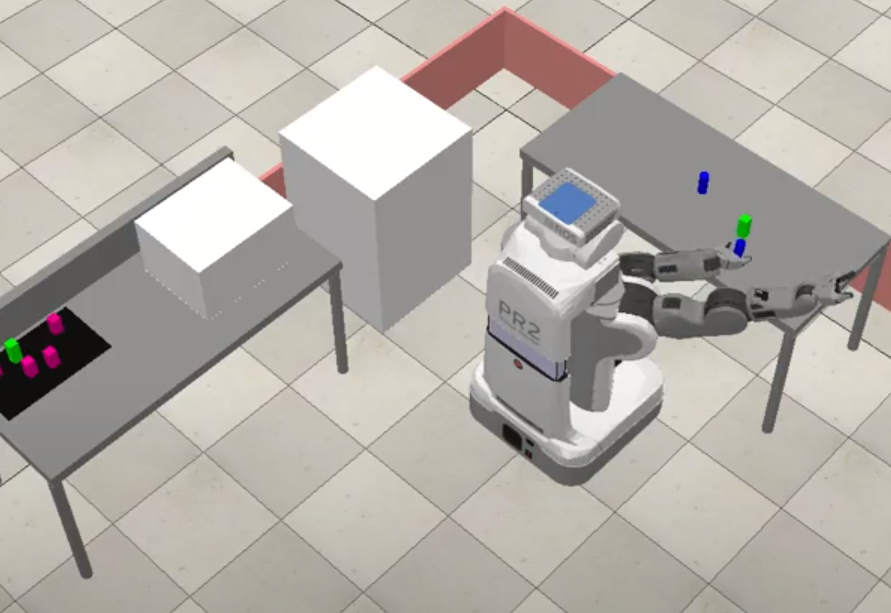}
}\\
\caption{
Simulation of the \textit{Habitat} experiment benchmark.
Samples (green blocks) must be sterilised and incubated, while glassware (blue) is cleaned and placed on the experiment bench.
Containers (pink) act as obstacles requiring non-monotonic rearrangement.
Snapshots (a)-(l) show various phases of the whole process.
}
\label{fig:habitat}
\end{figure*}

\subsection{\textsf{BENCHMARK\_2}: Habitat}
\label{sec:benchmark_habitat}

The second benchmark evaluates \textsf{TMP-EAOG} in a domain requiring non-monotonic and non-geometric actions, reflecting possible activities of a service robot performing scientific experiments inside a space habitat.
Unlike the previous benchmark, here the robot must combine symbolic state changes (for example, \textit{instrument cleaned}) with geometric manipulation and navigation. 
This scenario illustrates how autonomy could support astronauts in daily laboratory tasks, where limited crew time and confined workspaces require reliable robot assistance.

We model a situation where a PR2 mobile manipulator executes a simplified scientific experiment (see Figure \ref{fig:habitat}). 
On a workbench, samples (green blocks) are initially obstructed by containers (pink blocks). 
The robot must temporarily move the containers aside, retrieve the samples, clean them by placing them in a steriliser (large white cube), then activate a symbolic \texttt{wait} action representing sterilisation. 
Subsequently, samples are placed into a heating unit (small white cube) for incubation. 
Meanwhile, laboratory glassware (blue blocks) is collected from a side table, cleaned in the steriliser, and finally placed on the experiment bench. 
The procedure ends when both samples and glassware are properly prepared and arranged, representing the completion of a self-contained experiment run.

This benchmark highlights several challenges. 
\begin{itemize}
\item
\textit{Non-monotonicity}:
containers must be moved aside and later restored to their original location, analogous to temporary reconfiguration of habitat equipment.
\item
\textit{Non-geometric actions}: 
operations such as \texttt{sterilise} and \texttt{incubate} are symbolic actions with no \textit{geometric} effect, yet critical for task correctness.
\item
\textit{Multi-agent execution}: 
planning involves PR2’s two arms for manipulation and its base for navigation across different habitat areas.
\end{itemize}


Table~\ref{tab:habitat} reports average computation times. 
As in \textsf{BENCHMARK\_1}, motion planning dominates computation, with the right arm requiring hundreds of attempts to find feasible trajectories in cluttered configurations. 
Navigation plans are comparatively lightweight, though frequent, due to the need to shuttle between distinct experiment areas. 
The results confirm that while task-level reasoning (graph expansion, search) remains negligible in cost, motion feasibility checks drive the bulk of computation, underscoring the importance of tight task–motion integration.

\begin{table}[t]
\centering
\scalebox{0.7}{
\begin{tabular}{l c c}
\hline
\textit{Module}              & \textit{Avg. time [s]} & \textit{Std. dev. [s]} \\ 
\hline
\hline
AND/OR Graph expansion       & 7.8                   & 0.5 \\
Graph Net Search             & 0.5                   & 0.0 \\
Motion Planner (right arm)   & 318.0                 & 9.5 \\
Motion Planner (left arm)    & 121.8                 & 21.0 \\
Motion Planner (base)        & 56.9                  & 0.0 \\
\hline
\end{tabular}}
\caption{
Computation times for selected modules of \textsf{TMP-EAOG} in the habitat experiment benchmark; motion planning dominates.}
\label{tab:habitat}
\end{table}

\section{Conclusions}
\label{sec:conclusions}

We present \textsf{TMP-EAOG}, a task and motion planning framework that encodes symbolic abstractions in expanding AND/OR graphs, and couples them with motion feasibility assessments performed in-the-loop.
The recursive graph expansion mechanism enables plans to adapt online to infeasible actions or newly perceived constraints, a property particularly relevant for space robotics where uncertainty, occlusions, and communication delays prevent constant human supervision.

Through two benchmarks, we show how \textsf{TMP-EAOG} can address core challenges of autonomous space operations. 
The \textit{Towers of Hanoi} benchmark captured the need for sequencing dexterous manipulations under dynamic reachability constraints, analogous to handling tools or components on a cluttered lander deck.
The \textit{Habitat} benchmark showed how symbolic and geometric reasoning can be integrated in mixed-mode operations, such as scientific experiment preparation, where symbolic actions interleave with manipulation and navigation.

Results across benchmarks confirm that task-level search remains lightweight compared to motion planning, and that the expansion of AND/OR graphs grows primarily due to infeasible geometric actions and re-planning. 
These properties underline the importance of frameworks that explicitly integrate symbolic reasoning with geometric feasibility, rather than treating them as separate layers.

Future directions include extending \textsf{TMP-EAOG} to multi-robot scenarios, joint operations with humans, and -- possibly -- real space-qualified platforms.

\end{document}